\documentclass{article}
\usepackage{spconf,amsmath,graphicx}
\usepackage{amssymb,bm}
\usepackage{textcomp}
\usepackage{hyperref}
\usepackage{url}
\usepackage[keeplastbox]{flushend}
\usepackage{balance}
\usepackage{subfigure}
\usepackage{xcolor}
\usepackage{pgf}
\usepackage{threeparttable}
\usepackage{booktabs}
\usepackage{cite}
\usepackage{tikz}
\usetikzlibrary{automata,arrows,positioning,chains,shapes.multipart,fit,petri,shapes,backgrounds,decorations.pathmorphing,calc,shapes.misc, arrows, decorations.markings}
\usepackage{ifthen}

\newcommand{\eg}{e.\,g.,\ }
\newcommand{\ie}{i.\,e.,\ }

\newcommand{\et}{{et al.}}
\newcommand{\mb}{\mathbf}
\newcommand{\mc}{\mathcal}

\title{Attention-Augmented End-to-End Multi-Task Learning\\for Emotion Prediction from Speech}
\name{Zixing Zhang$^\star$$^{1,2}$, Bingwen Wu$^\star$$^3$, Bj\"orn Schuller$^{1,2,4}$
\thanks{Both authors contributed equally to this work.  This work was supported by the TransAtlantic Platform ``Digging into Data'' collaboration grant (ACLEW: Analysing Child Language Experiences Around The World), with the support of the UK's Economic \& Social Research Council through the research Grant No.~HJ-253479.}}
\address{$^1$GLAM -- Group on Language, Audio \& Music, Imperial College London, UK \\
  $^2$ audEERING GmbH, Germany \\ 
  $^3$ Department of Computing, Imperial College London, UK \\
  $^4$ ZD.B Chair of Embedded Intelligence for Health Care and Wellbeing,\\ University of Augsburg, Germany \\  
  {\small \tt \{zixing.zhang|bingwen.wu17\}@imperial.ac.uk}}

\begin{document}
\ninept
\maketitle
\begin{abstract}
Despite the increasing research interest in end-to-end learning systems for speech emotion recognition, conventional systems either suffer from the overfitting due in part to the limited training data, or do not explicitly consider the different contributions of automatically learnt representations for a specific task. In this contribution, we propose a novel end-to-end framework which is enhanced by learning other auxiliary tasks and an attention mechanism. That is, 
we jointly train an end-to-end network with several different but related emotion prediction tasks, \ie arousal, valence, and dominance predictions, to extract more robust representations shared among various tasks than traditional systems with the hope that it is able to relieve the overfitting problem. Meanwhile, an attention layer is implemented on top of the layers for each task, with the aim to capture the contribution distribution of different segment parts for each individual task. To evaluate the effectiveness of the proposed system, we conducted a set of experiments on the widely used database IEMOCAP. The empirical results show that the proposed systems significantly outperform corresponding baseline systems.  
\end{abstract}

\section{Introduction}
\label{sec:introduction}
Automatic speech emotion prediction endows machines with the capability of natural and empathic communication with humans, which is considered to be essential to sustain long-term human--machine interactions. In spite of remarkable advances over the past decades~\cite{Zhang18-Dynamic}, the extraction of representative features associated with emotions remains an open challenge. 
The conventional approaches normally extract a variety of acoustic descriptors, such as pitch and energy, on the frame level in the first place. Then, mostly they derive the super-segmental features via applying some mathematical functionals (\eg mean and maximum)~\cite{Eyben15-Real,Han14-Speech}, or counting the normalised occurrence frequency of certain frame-level acoustic feature units~\cite{Han18-Bags}.  

However, these approaches have several disadvantages. All these approaches largely require acoustic experts and psychologists to manually design the features. Only the feature attributes that explicitly showed high correlation with emotion, normally through extensive and carefully prepared experiments, will be selected~\cite{Eyben15-Real}, which is quite time-consuming and exhausting. Moreover, the effectiveness of selected features still heavily depends on the implemented pattern recognition model~\cite{Eyben15-Real}, resulting in their lower generality. In this regard, end-to-end learning has emerged as a promising alternative~\cite{Trigeorgis16-Adieu, Tzirakis17-End, Sarma18-Emotion, Huang18-End}. It aims to {\em automatically} explore the most salient representations related to the task of interest by using neural networks to jointly train the representation extraction process and the pattern recognition process,  wiping away the brute-force feature designing procedure. 

Since the inception~\cite{Graves14-Towards}, a number of end-to-end learning frameworks have been quickly and widely applied to various speech-related tasks, for example, speech recognition~\cite{Miao15-Eesen,Amodei16-Deep}, speaker recognition~\cite{Heigold16-End}, and speech synthesiser~\cite{Wang17-Tacotron}. 
As to speech emotion prediction, the first end-to-end work was shown in~\cite{Trigeorgis16-Adieu}, where the authors intended to extract implicit representations directly from digital raw signals by using one-dimensional Convolutional Neural Networks (CNNs) followed by Long Short-Term Memory (LSTM) Recurrent Neural Networks (RNNs) for learning a sequential pattern. Due to its great success, this end-to-end framework has been extended to deal with other modalities for emotion recognition, such as video and electroencephalogram signals~\cite{Tzirakis17-End, Huang18-End}. 
Likewise, other similar end-to-end frameworks were further proposed and investigated for speech emotion prediction. For example, in~\cite{Huang18-End}, the conventional CNNs were replaced by time-delay neural networks whereas the LSTM-RNNs were kept following behind.   

Nevertheless, these proposed frameworks easily suffer from overfitting~\cite{Trigeorgis16-Adieu}, leading to severe performance degradation when the data mismatch increases between the training and evaluation phases. This is because of not only the limited size of training data, but also further concerns, such as the task-specific training~\cite{Ruder17-overview}. 
In this regard, in this paper, we propose to integrate multi-task learning (MTL) into the end-to-end framework, which intends to jointly train several different, but related tasks simultaneously. By doing this, it is assumed that 
the more tasks are learnt simultaneously, the more common representations shared by all of the tasks and the less chance of overfitting on the original task will be gained~\cite{Baxter97-Bayesian}.

Moreover, when modelling the automated learnt representative sequence, the representations at each time point are normally equally considered~\cite{Trigeorgis16-Adieu,Tzirakis17-End}. This process largely ignores the different importance of the parts within one unit of analysis with respect to different emotions. For instance, the work done in~\cite{Mirsamadi17-Automatic} has shown that the short silence periods within an utterance often have little relevance with emotions. 
To this end, we further propose to implement an attention mechanism to the end-to-end frameworks, hoping that it can automatically learn the most interesting parts of an utterance containing strong characteristics relating to the given emotions.

Therefore, the main contribution of this paper pertains to the proposal of a novel end-to-end framework, which is augmented with an attention mechanism and jointly trained with multiple auxiliary tasks, for speech emotion prediction. To the best of our knowledge, this is the first time to investigate such an end-to-end framework in the context of speech processing.

\vspace{-.2cm}
\section{Related Work}
\label{sec:relateWork}
For speech emotion prediction, MTL has been frequently utilised. Eyben \et~\cite{Eyben12-multitask} firstly proposed to jointly train five different emotional dimensions for continuous emotion recognition. The experimental results have clearly indicated that the MTL model remarkably outperforms single-task-based models. Following this work, Han \et~\cite{Han17-From} combined the emotion prediction with an annotation uncertainty as joint tasks to be learnt together. Xia and Liu~\cite{Xia17-multi} suggested incorporating the losses from both the categorical and the dimensional emotion recognition to optimise the neural networks. Zhang~\et~\cite{Zhang17-Cross} investigated MTL in a cross-corpus scenario, where  many auxiliary tasks, such as corpus, domain, and gender distinctions, were considered to be optimised along with emotion recognition. Other similar works have also been done in~\cite{Parthasarathy17-Jointly}. However, most of these studies have focused on the usage of hand-crafted features. 

As to the attention mechanism, Mirsamadi~\cite{Mirsamadi17-Automatic} firstly integrated an attention layer within Deep Neural Networks (DNNs), resulting in a significant performance improvement for speech emotion prediction. Similarly, Zhao~\et~\cite{Zhao18-Exploring} implemented an attention layer right after the RNNs to extract the most interesting acoustic parts in the continuum. Apart from the RNNs and DNNs, the attention layer was also integrated with CNNs~\cite{Chen18-3D, Li18-attention}. All these works, nevertheless, were conducted under the usage of traditional hand-crafted features, and have not explicitly investigated the differences of attention in an MTL framework. 

\vspace{-.2cm}
\section{Attention-Augmented End-to-End Multi-Task Learning}
\label{sec:method}
\begin{figure}
    \centering
    \includegraphics[height=3.0in, width=3.5in, trim={0cm 0cm 4cm 0cm}, clip]{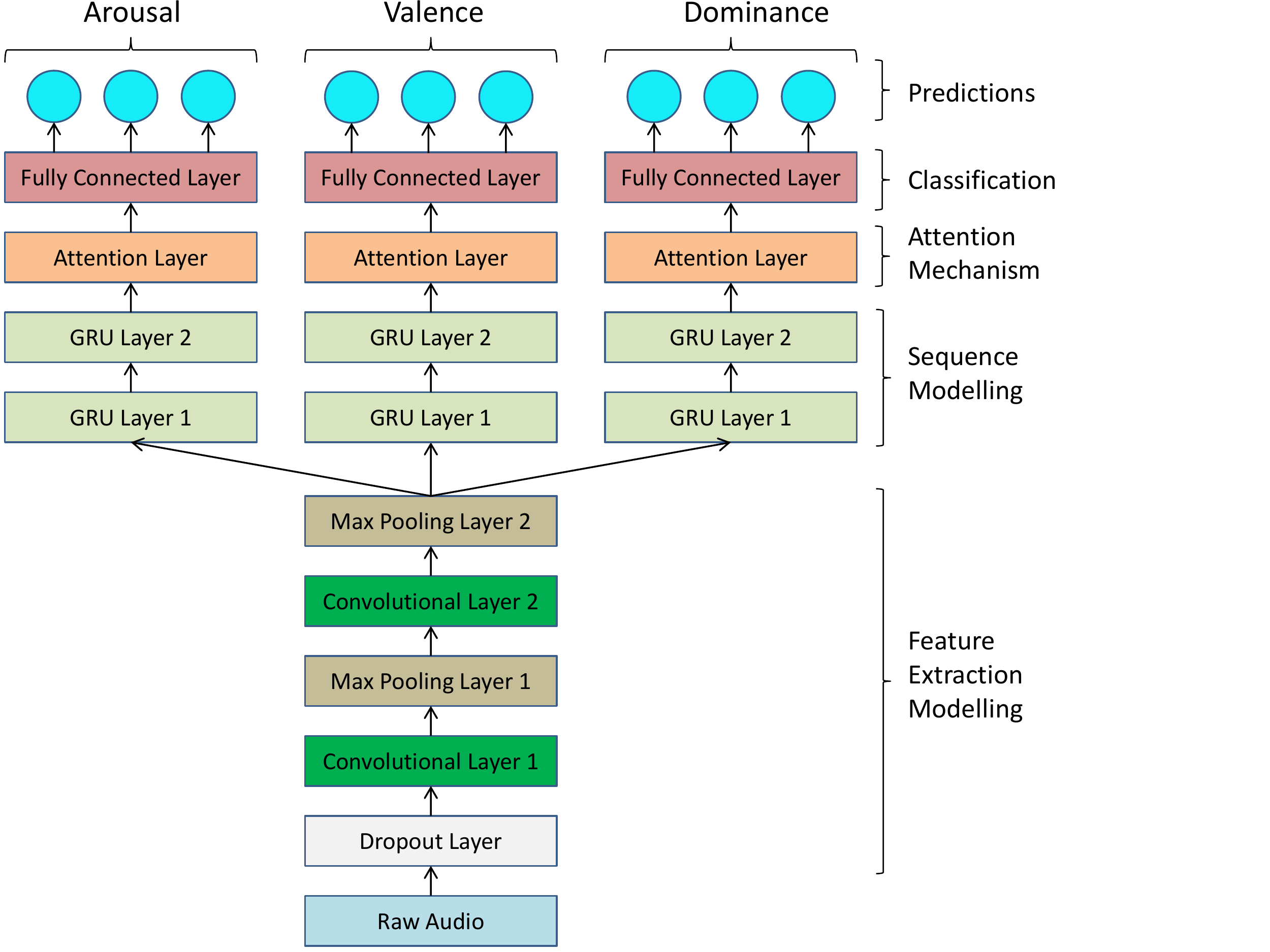}
    \caption{The framework of the attention-augmented end-to-end multi-task learning for speech emotion prediction.}
    \label{fig:e2e}
    \vspace{-.2cm}
\end{figure}

Figure~\ref{fig:e2e} illustrates the proposed end-to-end framework for speech emotion prediction, which can be considered as an extension of a basic end-to-end system, augmented with attention and MTL strategies. 
In the following subsections, we comprehensively describe the framework. 

\vspace{-.2cm}
\subsection{Single-Task End-to-End Framework}
\label{subsec:method_ste2e}
Despite several existing end-to-end frameworks for speech emotion recognition, we retained the basic network structure in our previous work~\cite{Trigeorgis16-Adieu,Tzirakis17-End}. This is due to its effectiveness being well demonstrated in continuous emotion recognition, and its widespread usage in many other computational paralinguistic tasks~\cite{Zhang18-Evolving}. 

The basic single-task-based end-to-end system generally consists of a feature extraction modelling and a sequence modelling. More specifically, the feature extraction modelling mainly consists of two one-dimensional convolutional layers each followed by an element-wise rectified linear non-linearity (ReLU) $max(0, x)$ and by additional max pooling layer.
The reason behind the usage of CNNs mainly lies in their well-known capability of feature extraction not only in image processing, but also in speech processing~\cite{Trigeorgis16-Adieu}. Particularly, one dropout layer is employed to increase the model generalisation. In contrast to the feature extraction modelling, the sequence modelling employs two recurrent layers equipped with Gated Recurrent Units (GRUs), due to their effectiveness in modelling temporal patterns and less complexity in comparison with LSTMs~\cite{Zhang17-Audio}.    

Given an utterance in form of raw audio signals $s(t)$, it is firstly split into several sequential segments $\{\mb{s}_1, \ldots, \mb{s}_L\}$ ($L$ indicates the number of the obtained segments given an utterance) with a sliding fixed-length window. Then, each segment is successively fed into the feature extraction modelling ($f_c$) so that input raw signals of each segment are transformed into one vector (\ie representation). That is, given a segment $\mb{s}_i$, one obtains 
\begin{equation}
    \mb{v}_i = f_c(\mb{s}_i),
\end{equation}
which is assumed to well represent the temporal speech patterns. After that, the extracted representation {$\mb{v}_i$} is further successively fed into the sequence modelling ($f_r$), \ie
\begin{equation}
    \mb{h}_i = f_r(\mb{v}_i),
\end{equation}
leading to a new output sequence from the last recurrent layer, \ie $\{\mb{h}_1, \mb{h}_2, \ldots, \mb{h}_L\}$.  Normally, only the last output of the sequence $\mb{h}_L$ is used and fed into the final softmax layer for emotion prediction. 

\vspace{-.2cm}
\subsection{Weighted Pooling with Attention}
The principle of the attention mechanism originally stemmed from the characteristic of human perception. That is, humans normally focus attention selectively on parts of the visual or auditory space to acquire information when and where it is needed, and combine information from different fixations over time to build up an internal representation of the scene~\cite{Mnih14-Recurrent}. Nowadays, the attention mechanism has been widely used in image processing and natural language processing~\cite{BahdanauCB15-Neural}. 

By far, a variety of attention mechanisms have been investigated in machine learning. According to whether the calculation of attention requires to access positions across sequences, they can be generally categorised into inter-attention and intra-attention mechanisms. Intra-attention, also known as self-attention, is often used to compute a representation of a sequence by leveraging different importance of the parts in a sequence. In this contribution, we employed an intra-attention layer following the last recurrent layer as illustrated in the top of Fig.~\ref{fig:e2e}. 

Mathematically, given an output sequence of the last recurrent layer $\{\mb{h}_1, \mb{h}_2, \ldots, \mb{h}_L\}$, the attention layer tries to deliver an utterance-level representation for such a sequence with the following equation 
\begin{equation}
    \mb{r} = \sum_{i=1}^{L}\alpha_i \mb{h}_i, 
\end{equation}
where $\alpha_i$ stands for the weight of the output $\mb{h}_i$ at the $i$-th frame. From this equation, it can be seen that the attention layer can be considered as a weighted-average pooling layer, in comparison with the traditional maximum, average, or the last output pooling strategies. Finally, the obtained new representation $\mb{r}$ is fed into one fully connected layer for emotion prediction. 

Therefore, the calculation of weight $\alpha_i$ becomes the central problem of the attention layer. Specifically, $\alpha_i$ is computed by 
\begin{equation}
    \alpha_i = \frac{\text{exp}(\mb{w}^T\mb{h}_i)}{\sum_{t=1}^{L} \text{exp}(\mb{w}^T\mb{h}_t)},
\end{equation}
where $\mb{w}$ is the learnable parameter vector, and the inner product between $\mb{w}$ and $\mb{h}_i$ is interpreted as a score for the contribution of the frame $t$. Besides, in this equation, a softmax function is applied leading to the sum of the weight distribution to be a unity.

\vspace{-.2cm}
\subsection{Joint Training with Auxiliary Tasks}
As discussed in Section~\ref{sec:introduction}, end-to-end learning frameworks normally require massive training data, which, however, are largely absent in the context of emotion recognition, resulting in a severe overfitting problem. To improve the model generalisation, in this paper we endeavour to seek help from other auxiliary tasks through MTL. The underlying idea is that the model learns more tasks simultaneously will contribute to more robust learnt representations that capture all of the tasks~\cite{Ruder17-overview}. 

Figure~\ref{fig:e2e} illustrates the structure of the proposed MTL. From the technical view of point, MTL is a process of learning multiple tasks \textit{concurrently}. Typically, there is one main task and one or more auxiliary tasks. By attempting to model the auxiliary tasks together with the main task, the model learns shared information among tasks, which may be beneficial to learning the main task. 
In this paper, the tasks refer to three-dimensional emotions, \ie arousal, valence, and dominance predictions. 

Mathematically, the objective function in MTL can be formulated as:
\begin{equation}
 \mc{J}(\bm\theta_0)={\sum_{m=1}^{M}w_m L_m(\mb{x},{y}_{m},[\bm{\theta}_m;\bm{\theta}_c])+{\lambda}R(\bm{\theta}_0)},
\label{eq:mtl}
\end{equation}
where $M$ denotes the number of tasks and $L_m(\cdot)$ represents the loss function of the task $m$, which is weighted by $w_m$. The weights $w_m$ are optimised by a random search. 
$\bm{\theta}_c$ and $\bm{\theta}_m$ represent, respectively, the shared and task-specific model parameters with respect to the task $m$, whereas $\bm{\theta}_0$ indicates all shared and task-specific network parameters, and $\lambda$ is a hyper-parameter that controls the importance of the regularisation term $R(\bm{\theta}_0)$ (\ie L2 in our case). In the network training process, the network is optimised by minimising the objective function $\mc{J}(\bm\theta_0)$.

\vspace{-.2cm}
\section{Experiments and Results}
\label{sec:experiment}
In this section, we implement and evaluate our approach for emotion classification on an emotion database.

\vspace{-.2cm}
\subsection{Selected Database}
\label{subsec:database}
To validate the proposed paradigm, we used the widely used Interactive Emotional dyadic MOtion CAPture (IEMOCAP) database, which contains approximately 12 
hours of audio-visual recordings from five pairs of experienced actors~\cite{busso2008iemocap}. For each improvised interaction between two actors, they communicated with each other in scenarios where specific emotions were elicited.
The recordings were then segmented into utterances and further annotated in all three-dimensional aspects, \ie activation, valence, and dominance, on a five-point scale by at least two different annotators. 

For our experiments, only the audio recordings were utilised. Following the work of~\cite{Metallinou12-Context}, we further discretised the five-point scale into three levels (classes) -- low level contains ratings in the range [1,2], middle level contains ratings in the range (2,4), and high level contains ratings in the range [4,5]~\cite{Metallinou12-Context}. 
We divided the dataset into three speaker independent partitions, \ie $6\,319$ for the training set (session 1-3), $1\,811$ for the development set (session 4), and $1\,819$ for the test set (session 5). All the recordings were sampled with 16 kHz.  


\vspace{-.2cm}
\subsection{Implementation Details}
\label{subsec:setup}
Before feeding the raw speech signal into the network, we applied an online standardisation to the development and test sets by using the mean and standard deviation information from the training set. The raw speech signals were then split into sub-segments with a fixed-size window of 40\,ms at a  step size of 10\,ms. Given the 16\,kHz sampling rate of raw signals, the network input vector is of dimension 640 for each sub-segment. 
For the first and second convolutional layers, we used 40 filters with the size of 40, resulting in 40 feature maps after each layer. For the first max pooling layer, we took a kernel with the size of two in a zero-padding strategy, leading to feature maps with a dimension of 320; whereas, for the second max pooling layer, we used a cross-channel max pooling with the pool size of 10, yielding to four feature maps with the dimension of 320.
Finally, the obtained feature map is expanded and concatenated as a long vector with 1\,280 dimensions, which is the extracted representation for each sub-segment. To improve the model generalisation, we set the keep probability of the dropout layer to be 0.9. For the sequence modelling, we employed 128 nodes per GRU hidden layer. The training of the proposed framework was conducted using the Adam optimisation algorithm with a learning rate of 0.0001. Note that all these network and training hyper-parameters were optimised on the development. 

To evaluate the model performance, we utilised the frequently used metric Unweighted Average Recall (UAR), \ie the sum of classwise recall divided by the number of classes, for emotion recognition. 

\vspace{-.2cm}
\subsection{Results and Discussion}
\label{subsec:results}

\begin{table}[!t]
\centering
\caption{Performance comparison (UAR: unweighted average recall) between the proposed attention-augmented end-to-end multitask learning system with other baseline systems as well as other traditional recognition models on the development and the test partitions for activation, valence, and dominance predictions. OS: \  openSMILE features; SVMs: support vector machines; RNNs: recurrent neural networks; STL: single-task learning; MTL: multi-task learning; e2e: end-to-end learning; att.: attention. The sign of `$\star$' indicates statistic significance (one-tailed $z$-test, $p<.05$) of performance improvement of the proposed systems in comparison with the baseline system (\ie e2e STL).}
\label{tab:results}
\vspace{0.2cm}
\begin{threeparttable}
  \begin{tabular}{lllllll}
  \toprule
 UAR [\%] & \multicolumn{2}{c}{arousal} & \multicolumn{2}{c}{valence} & \multicolumn{2}{c}{dominance} \\
 methods & dev  & test & dev  & test & dev   & test  \\
  \midrule 
  OS+SVMs & 52.1 &	50.5 &	50.5 &	49.8 & 39.1 & 48.7  \\
  OS+RNNs & 57.8 &	53.1 &	51.8 &	51.0 & 56.4 & 49.6  \\
  \midrule 
    e2e STL & 45.3 & 45.1 & 60.9 & 60.1 & 50.9  & 51.1  \\
    e2e STL+att. & 46.5 & 45.4 & 61.4 & 60.7 & 51.4  & 52.6  \\
    e2e MTL & 46.2 & 44.0   & 64.6 & 63.4 & 52.3  & {\bf 53.9}  \\
    e2e MTL+att. & {\bf 48.7}$^\star$ & {\bf 48.5}$^\star$ & {\bf 66.2}$^\star$ & {\bf 63.8}$^\star$ & {\bf 53.4}  & 51.6  \\
  \bottomrule
  \end{tabular}
  \vspace{-.2cm}
\end{threeparttable}
\end{table}

\begin{figure}
    \centering
    \vspace{-.2cm}
    \includegraphics[height=3.in,width=3.2in, trim={0cm 0cm 11cm 4cm}, clip]{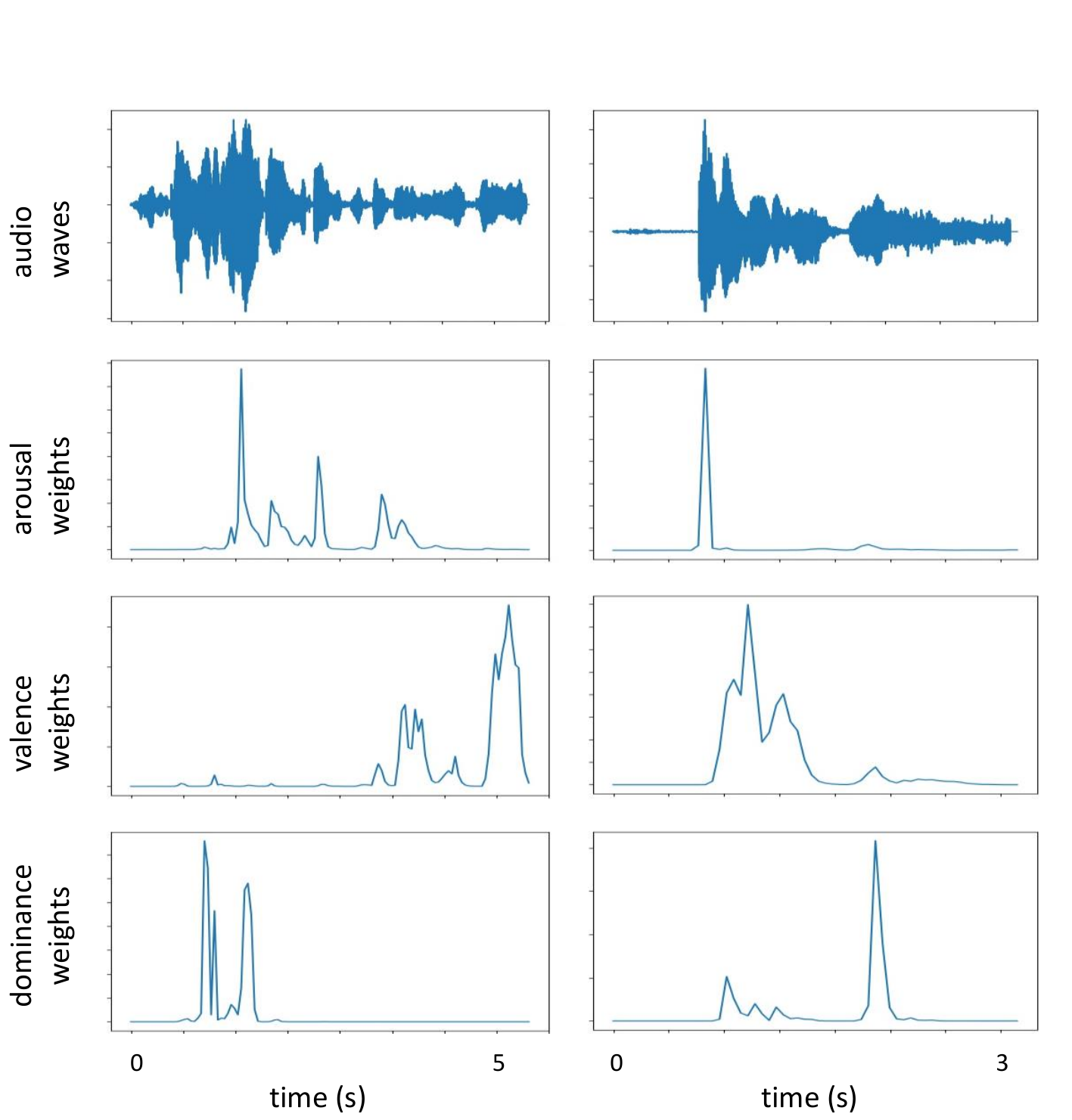}
    \vspace{-.2cm}
    \caption{Automatically learnt attention distribution for arousal (second row from top), valence (third row), and dominance (fourth row) predictions for two randomly selected audio wave files (first row).}
    \label{fig:attention}
    \vspace{-.2cm}
\end{figure}

To compare the performance of the proposed approach with other traditional speech emotion prediction systems, we conducted two experiments with hand-engineered acoustic features. That is, we used our opensource toolkit {openSMILE}~\cite{Eyben15-Real} to extract a minimalistic expert-knowledge based feature set~\cite{Eyben16-TGM}, which contains 23 Low-Level Descriptors (LLDs). After that, we applied a set of statistical functionals to the LLDs, leading to 88 acoustic features (\ie eGeMAPS) on the utterance level. As to the classifier, we utilised the sequence classifier of RNNs to model  frame-level features; whereas utilised the static classifier of Support Vector Machines (SVMs) to model the utterance-level features. Both systems have been successfully and frequently utilised for speech emotion prediction~\cite{Eyben16-TGM, Han14-Speech}. 

Table~\ref{tab:results} shows the obtained results in terms of UAR from the proposed attention-augmented end-to-end MTL system, the related baseline systems, as well as the aforementioned other state-of-the-art systems. It is noted that the basic end-to-end (e2e) learning systems refer to the ones without attention and MTL learning strategies (see Section~\ref{subsec:method_ste2e}), and take the last output from the last recurrent layer for a final prediction. 
It can be seen that the e2e systems are competitive to the two state-of-of-the-art systems based on hand-crafted features for both the valence and dominance predictions but not for the arousal prediction. This generally confirms our previous findings~\cite{Trigeorgis16-Adieu,Tzirakis17-End}. 

When integrating the attention strategy into the baseline systems (\ie e2e STL), one can note that the system performance is generally improved on all three prediction tasks. These findings suggest that the attention mechanism does not only work in the conventional learning framework with hand-crafted features~\cite{Han14-Speech}, but also in the proposed end-to-end framework.  
In parallel, when conducting the MTL strategy into the baseline systems (\ie e2e STL), similar observations are made. That is, the end-to-end MTL systems are superior to the task-specific baseline systems in most cases. This conclusion implies that the MTL method can partially increase the generalisation of the extracted representations, \ie the information learnt from other auxiliary tasks can benefit the task of interest, even in an end-to-end learning framework. 

Moreover, the incorporation of attention and MTL strategies achieves the best performance in five out of six cases. For example, the obtained results for arousal and valence predictions are achieved at 48.5\,\% and 63.8\,\% UAR, which significantly (one-tailed $z$-test, $p<.05$)  outperform the baseline results (\ie 45.1\,\% and 60.9\,\% UAR) on the test set. This suggests that both attention mechanism and MTL can work in a complementary way.  

To further investigate the effectiveness of the attention mechanism in the proposed end-to-end MTL framework, we illustrate the learnt attention across different tasks for two audio files in Fig.~\ref{fig:attention}. Generally speaking, one can observe that the learnt attention weight distributions remarkably differ each task. That is, the arousal, valence, and dominance prediction tasks learnt their individual higher attention on the same segment parts, which matches our previous assumption in Section~\ref{sec:introduction}. Particularly, one can see that for arousal prediction the learnt attention weights (refer to the second row of Fig.~\ref{fig:attention}) are highly correlated with the parts with high amplitude. Nevertheless, a similar observation is not made for valence prediction (refer to the third row of Fig.~\ref{fig:attention}). In contrast, the parts with low speech amplitude often contribute more than the parts with high amplitude. This matches our previous knowledge that the valence prediction has limited relation to speech amplitude~\cite{Eyben15-Real}. In addition, from the fourth row of Fig.~\ref{fig:attention}, it can be seen that dominance prediction lays more attention on some parts with high amplitude.  

\vspace{-.1cm}
\section{Conclusion}
\label{sec:conclusion}
With an end-to-end (e2e) learning framework, we, on the one hand, took a multi-task learning (MTL) strategy to improve the robustness of the learnt representations that are shared among several tasks. On the other hand, we integrated a self-attention layer on top of the layers for each prediction task, in order to distil more salient representations on the utterance level for a task of interest. 
 
  The experimental results obtained by performing experiments on the IEMOCAP database have shown that either the MTL-based or the attention-augmented e2e systems outperform the single-task-based e2e systems, which suggests the effectiveness of the proposed e2e learning framework. However, we also find that for arousal the introduced frameworks are inferior to the baseline systems with the classic functional-based features. This might be because the hand-crafted features are somewhat able to better represent the patterns for arousal due to its simplicity than for other tasks (\ie valence), according to our prior knowledge.  
  
 


\newpage\pagebreak
\bibliographystyle{IEEEtran}
\bibliography{refs}

\end{document}